\documentclass[10pt,twocolumn,letterpaper]{article}

\usepackage{utils/iccv}   
\usepackage{cuted}
\usepackage{multirow}
\usepackage{colortbl}
\usepackage{gensymb}

\definecolor{mygray}{rgb}{0.5,0.4,0.2}

\newcommand{\ModelName}{LongTextAR}

\usepackage{bbm}

\usepackage[most]{tcolorbox}
\usepackage{listings}

\usepackage{multicol}

\definecolor{customblue}{rgb}{0.2, 0.3, 0.8}
\definecolor{customgreen}{rgb}{0.1, 0.6, 0.3}

\newcolumntype{x}[1]{>{\centering\arraybackslash}p{#1pt}}
\newcolumntype{y}[1]{>{\raggedright\arraybackslash}p{#1pt}}
\newcolumntype{z}[1]{>{\raggedleft\arraybackslash}p{#1pt}}

\usepackage{soul}  % Provides \ul{}

\definecolor{iccvblue}{rgb}{0.21,0.49,0.74}
\usepackage[pagebackref,breaklinks,colorlinks,allcolors=iccvblue]{hyperref}

\title{Beyond Words: Advancing Long-Text Image Generation via Multimodal Autoregressive Models}

\author{Alex Jinpeng Wang$^{1}$ \quad
Linjie Li$^{2}$  \quad
Zhengyuan Yang$^{2}$  \quad
Lijuan Wang$^{2}$ \quad
Min Li$^{1}$ \\
$^1$Central South University \quad 
$^2$Microsoft \\
\url{https://fingerrec.github.io/longtextar}{}
}

\begin{document}
\maketitle
\begin{abstract}
Recent advancements in autoregressive and diffusion models have led to strong performance in image generation with short scene text words. 
However, generating coherent, long-form text in images, such as paragraphs in slides or documents, remains a major challenge for current generative models. 
We present the first work specifically focused on long text image generation, addressing a critical gap in existing text-to-image systems that typically handle only brief phrases or single sentences.
Through comprehensive analysis of state-of-the-art autoregressive generation models, we identify the image tokenizer as a critical bottleneck in text generating quality. 
To address this, we introduce a novel text-focused, binary tokenizer optimized for capturing detailed scene text features. 
Leveraging our tokenizer, we develop \ModelName, a multimodal autoregressive model that excels in generating high-quality long-text images with unprecedented fidelity. 
Our model offers robust controllability, enabling customization of text properties such as font style, size, color, and alignment.
Extensive experiments demonstrate that \ModelName~significantly outperforms SD3.5 Large~\cite{sd3} and GPT4o~\cite{gpt4o} with DALL-E 3~\cite{dalle3} in generating long text accurately, consistently, and flexibly. 
Beyond its technical achievements, \ModelName~opens up exciting opportunities for innovative applications like interleaved document and PowerPoint generation, establishing a new frontier in long-text image generating.
\end{abstract}    
\section{Introduction}
\label{sec:intro}
%paragraph1: introduce the history of text rendering and then multi-modality auto-regressive model.
Text is an integral part of our visual environment, appearing  in diverse formats such as logs, banners, book covers, and newspapers. 
In recent years, there has been growing interest in rendering text  as images and processing it in pixel space, leading to significant advancements in both natural-language processing (NLP)~\cite{PIXEL,pixar,pix2struct,ptp} and multi-modality model~\cite{clippo,leveraging,mplug}.
However, generating images that contain accurate, aesthetically pleasing, and contextually coherent text is a challenging task.

% Traditional research in visual text rendering has seen some progress, with approaches demonstrating the benefits of powerful large language models as text encoders~\cite{textdiffuser,glyphcontrol,ca_model} or fine-grained guidance like layout~\cite{anytext}. 
% For example, GlyphControl~\cite{glyphcontrol} enables positioning through specified guidelines, while TextDiffuser~\cite{textdiffuser} employs character-level segmentation masks. Later advancements like TextDiffuser2~\cite{textdiffuser2} use large language models to reduce the need for manual prompt engineering. 
% While these methods have improved text rendering accuracy, they exhibit several key limitations:
% 1. \textit{Heavy reliance on specific layout generation or manual keyword specification}.
% 2. \textit{Limited capability to render long text, typically restricted to single short sentence}.

\begin{figure}
    \centering
    \includegraphics[width=\linewidth]{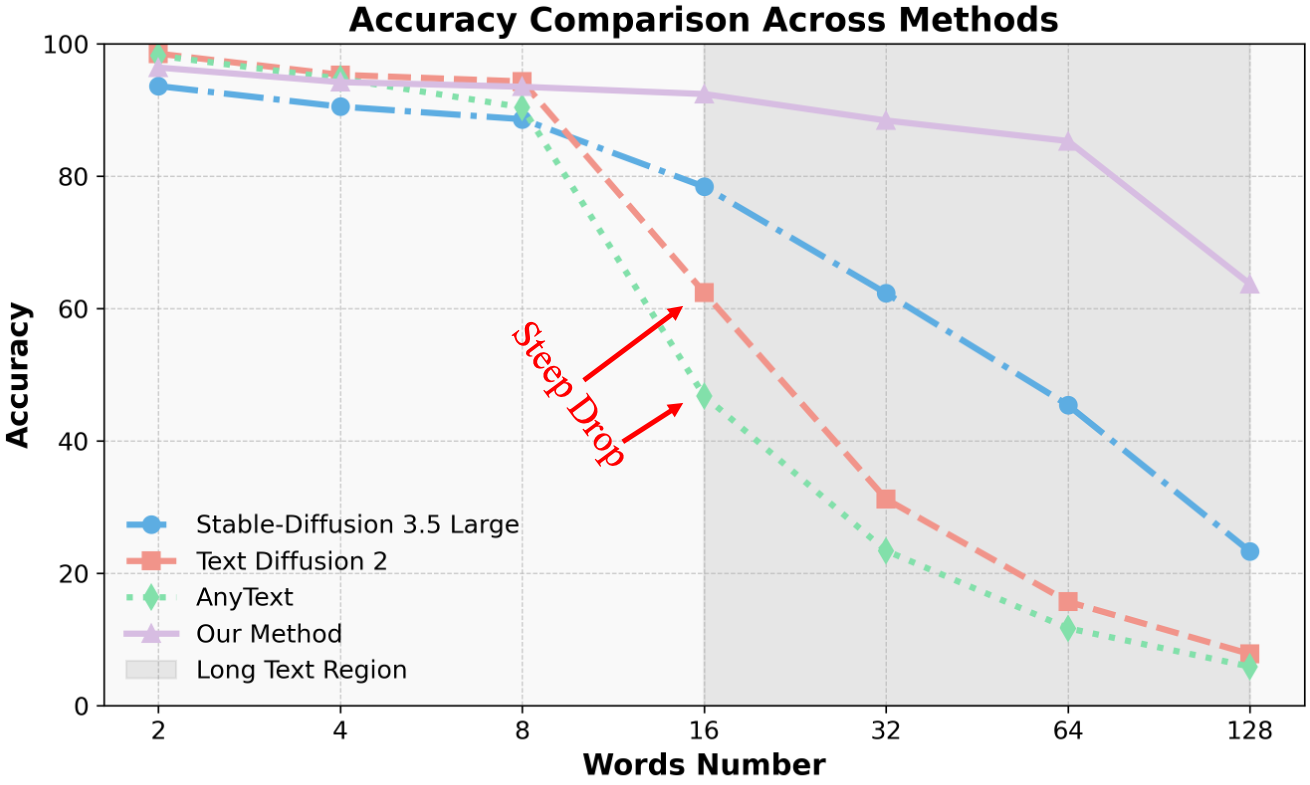}
    % \vspace{-2em}
   \caption{ \textbf{Breaking the Limits: Long-Text Image Generation Remains Elusive for Existing Models.} 
   State-of-the-art text rendering models, such as Text Diffusion 2~\cite{textdiffuser2} and AnyText~\cite{anytext}, perform well on short text but struggle with longer passages. 
   Large diffusion models like Stable Diffusion 3.5 Large~\cite{sd3} can handle longer text but exhibit lower accuracy. The text recognition on generated images was conducted using Qwen2-VL~\cite{qwen2_vl} model. 
   For this evaluation, we sampled 140 examples from the interleaved Obelics~\cite{obelics} dataset with truncation. }
    \label{fig:word_acc}
\end{figure}

Existing approaches in visual text generation have progressed with methods leveraging large language models as text encoders~\cite{textdiffuser, glyphcontrol, ca_model} and fine-grained guidance, such as layout control~\cite{anytext}. GlyphControl~\cite{glyphcontrol} enables positioning through specified guidelines, while TextDiffuser~\cite{textdiffuser} employs character-level segmentation masks. More recent advances, like TextDiffuser2~\cite{textdiffuser2}, use language models to reduce manual prompt engineering. 
Despite these improvements, existing methods face a shared key limitation: \textit{Limited capability to render long text, typically restricted to single short sentence}, as shown in Figure~\ref{fig:word_acc}.

The rise of large diffusion models~\cite{sd3,dalle3}  has begun to address this limitation.
As illustrated in Figure~\ref{fig:word_acc}, these models have expanded their capacity to \textbf{handle longer textual prompts up to tens of words, but they still face a fundamental limitation: the context window}. 
While diffusion models like SD-XL3~\cite{sd3}, MidJourney~\cite{midjourney}, and FLUX~\cite{flux} have made impressive strides, they support relatively short context lengths, typically only up to 77 tokens in the case of SD3.5 Large~\cite{sd3}. 
This restriction makes them less suited for handling more extensive real-world text, where the accuracy of image generation deteriorates significantly as the length of the input increases.
Despite advancements in document-level text perception~\cite{gpt4o,internvl,deepseek,qwen2_vl}, where image-to-text models excel, \textbf{the reverse task of generating high-quality images from long text descriptions remains underdeveloped, highlighting a significant gap}.

In contrast, autoregressive (AR) models based on large language models (LLMs), such as LlamaGen~\cite{llamagen} and Chameleon~\cite{chameleon}, offer a promising alternative. 
\textbf{These AR models can process much longer sequences of text—up to 4096 tokens} in Chameleon~\cite{chameleon} and 2048 tokens in LlamaGen~\cite{llamagen}—while achieving comparable text-to-image generation performance to diffusion models. 
However, AR models still struggle with precise text rendering, particularly in the context of complex or lengthy text inputs~\cite{chameleon}, highlighting the need for further improvement.

Given these observations, we analyze the Vector Quantization representation in AR models, identifying codebook embedding limitations as a key bottleneck for text rendering. 
To address this, we develop TextBinarizer, a specialized text-focused tokenizer that delivers clearer, more consistent reconstructions for complex documents with long text. 
We present \ModelName, a multi-modality AR model designed for advanced text rendering that generates high-quality text images across various styles. 
Our experiments show that \ModelName~ produces clearly readable text and outperforms state-of-the-art models including SD3.5 Large~\cite{sd3} and GPT-4o~\cite{gpt4o} with Dall-E3~\cite{dalle3}.

\ModelName~ excels in controllable text rendering, supporting customizations in font type, size, color, and alignment. 
Unlike SD3.5 Large and DallE3, which often struggle with specific instructions, \ModelName~ consistently produces accurate results under various constraints. 
We also co-trained \ModelName~ on both synthetic long-text image rendering and natural image generation tasks, finding that our model performs well on natural image generation despite its focus on synthetic text image data—demonstrating the versatility of our approach.

The main contribution includes:
\emph{i. \textbf{First Long-Text Image Generation Model:}} We present the first model specifically designed for long-text image generation, addressing a significant gap in existing text-to-image methods that typically handle only short sentences.
\emph{ii. \textbf{Identifying Tokenization Bottlenecks}:} We pinpoint weak tokenization as a critical barrier for effective text rendering in existing multimodal autoregressive models, such as Chameleon.
\emph{iii. \textbf{Developing an Text-focus Tokenizer:}} We design a binary tokenizer tailored for text rendering, achieving significantly better reconstruction quality on complex documents compared to traditional VQ-based models. 
\emph{iv. \textbf{Versatile Text Rendering:}} \ModelName~ offers customizable text rendering with control over font attributes while generalizing to natural image generation through co-training. Our experiments demonstrate its potential for applications like document generation and PowerPoint editing.
\section{Related Works}
\label{sec:related_works}

\paragraph{Visual Text Rendering.}
Text is omnipresent in daily life, appearing in  posters, TV, road signs, and more. 
Given its visual nature, it is natural to treat text processing as an image-based task~\cite{clippo,pix2struct}. PTP~\cite{ptp} and PIXEL~\cite{PIXEL} have explored this approach in Natural Language Processing. 
In computer vision, some works~\cite{anytext,glyphcontrol,glyphdraw,textdiffuser,pixart,ca_aware} try to render text into image due to enhanced rendering could significantly benefit text-centric design tasks like logos and posters. 
Along this line, traditional methods~\cite{glyphdraw, textdiffuser,textdiffuser2} use layout guides.
Concurrent work DnD Transformer~\cite{dnd_transformer} focusing on unconditioned long-text image generation and works well on longer text data like PDF.
%while TextDiffuser-2~\cite{textdiffuser2} integrates large language models (LLMs) to improve layout and spelling. 
However, these approaches struggle with rendering longer text, limiting diversity and precision.
Large diffusion models like SD-XL3~\cite{sd3} and DALL·E 3~\cite{dalle3} can render longer text but lack font control and support limited token lengths. 
In contrast, we focuses on LLM-based Autregressive (AR) models that handle much longer sequences and introduce controllable font variations, enhancing diversity for text-centric designs.

\paragraph{Autoregressive Image Generation.}
Autoregressive image generation models~\cite{unifiedio2,emu,beit} models has emerged as a powerful alternative to diffusion-based methods~\cite{emu2,chameleon}, leveraging sequential token-based approaches. It relies on vector quantization~\cite{anygpt,bai2024sequential,gemini} to convert images into discrete tokens. Early works~\cite{ramesh2021zero,parti} introduced a two-stage pipeline: tokenize images using VQ-VAE~\cite{vq_vae} or VQ-GAN~\cite{vqgan}, then model the token sequences with an autoregressive transformer.
LlamaGen~\cite{llamagen} improved tokenization and adopted advanced language modeling, narrowing the gap with diffusion models.
Furthermore, models like Chameleon~\cite{chameleon}, SeedX~\cite{seedx}, Show-O~\cite{showo}, and Lumina-mGPT~\cite{lumina_mgpt} unified text and image sequences, fine-tuning on text-to-image pairs. 
In this work, we focus on integrating our text-focused tokenizer into an AR framework. By leveraging our diverse, text-rich dataset, we aim to advance text-image generation, delivering more accurate and stylistically varied outputs conditioned on long, detailed prompts.

\paragraph{Visual Tokenization in Multi-Modal Autoregressive Models.}
The first step in multi-modal autoregressive modeling is converting images into discrete token IDs. 
The vector quantization (VQ) is introduced into VQ-VAE~\cite{vq_vae}  to learn discrete representations of images.
VQ-GAN~\cite{vqgan} use larger resolution and propose new train targets. 
EMU~\cite{emu} and EMU2~\cite{emu2} use Casual Transformer to get 1D causal sequences.
SEED~\cite{seed} propose discrete image tokenizer with 1D causal dependency.
MAR~\cite{mar} propose to conduct diffusion autoregressive without VQ.
Most recent models like ViT-VQGA~\cite{vit_vqga}, Chameleon~\cite{chameleon}, Show-O~\cite{showo} and LlamaGen~\cite{llamagen} leverage the same tokenizer architecture from VQ-GAN~\cite{vqgan}.
However, a critical factor in VQ is the size of the codebook, VQ models often use relatively small codebooks~\cite{vqgan,dalle3}. 
MagVit-V2~\cite{magvit2,open-magvit2} and Infinity~\cite{infinity} fixes this flaw by employing binary quantization with an expanded vocabulary. 
Different from these advancements, our work is the first to specifically address long text rendering, pushing the boundaries of what's possible in this domain while introducing a specialized text-focused tokenizer.

\section{Advancing Tokenization for Text-Rich Images Generation}

Although Autoregressive (AR) models boast longer context windows than large diffusion models, they falter when generating images with dense textual content.
This section uncovers the fundamental bottleneck in existing tokenization strategies and introduces TextBinarizer, a novel approach that significantly enhances text preservation.

\subsection{Tokenization Challenges in Text Rendering}

AR multimodal models require text and images to be converted into a unified sequence of discrete tokens before generation. 
The dominant paradigm, VQ-based tokenization~\cite{llamagen,lumina_mgpt,chameleon}, constructs a shared vocabulary by incorporating both subword text tokens (e.g., BPE~\cite{sentencepiece}) and visual tokens from a VQ codebook. 
While this framework excels in generating natural images, it exhibits \textbf{catastrophic failures} when encoding fine-grained text details in images~\cite{vqgan}. 
Our investigation in Section~\ref{sec:tokenizer_impact} demonstrates that these limitations persist even with large-scale training data, revealing the tokenization process itself as a fundamental bottleneck.

Through extensive analysis of state-of-the-art VQ tokenizers from Taming Transformer~\cite{vqgan} and Chameleon~\cite{chameleon}, we identify inherent limitations in text rendering quality. 
The VQ architecture prioritizes global image structure over fine-grained character details, utilizing small codebooks that inadequately capture typographic variations. 
These architectural constraints lead to poor text reconstruction, which further degrades when generating text-heavy images through AR models, resulting in blurry or illegible output. 
To overcome these challenges, we introduce TextBinarizer, a specialized tokenization approach that preserves textual fidelity in image generation.

\begin{figure}
    \centering
    \includegraphics[width=\linewidth]{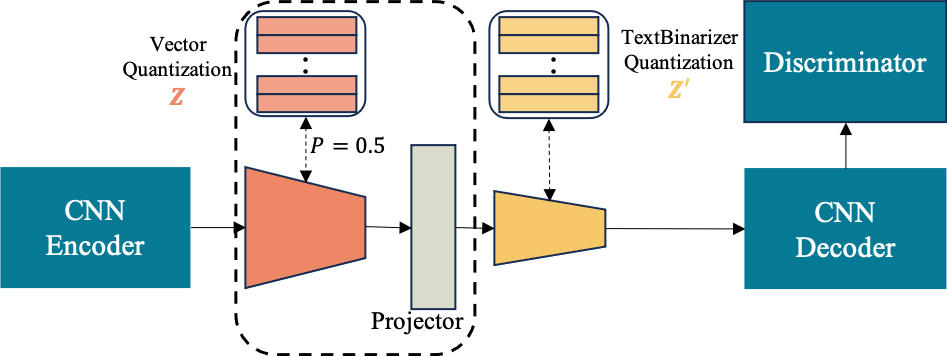}
    % \vspace{-1.5em}
    \caption{
    \textbf{TextBinarizer implementation details}.
    This approach allows for direct quantization.}
    \label{fig:text_binarizer}
\end{figure}

\subsection{TextBinarizer: A Text-Centric Tokenization}

\paragraph{Beyond Vector Quantization: A Bitwise Approach.}
TextBinarizer introduces a novel \textbf{bitwise quantization scheme} that encodes high-dimensional embeddings into binary tokens, surpassing traditional VQ methods in preserving textual details. 
Inspired by recent works~\cite{bsq,magvit2,infinity} demonstrating that binary representations excel at capturing fine-grained features with remarkable fidelity, we adopt a binary codebook approach specifically optimized for preserving detailed textual characteristics. 
This choice enables our model to maintain high-quality text rendering while remaining computationally efficient.
This architecture is illustrated in Figure~\ref{fig:text_binarizer}.
For an embedding vector \( \mathbf{x} \in \mathbb{R}^{\log_2 K} \), quantization occurs per dimension:
\begin{equation}
q(x_k) = \text{sign}(x_k) = -\mathbbm{1}\{x_k \leq 0\} + \mathbbm{1}\{x_k > 0\}.
\end{equation}

The final token index is computed as:
\begin{equation}
\text{Index}(\mathbf{x}) = \sum_{k=1}^{\log_2 K} 2^{k-1} \mathbbm{1}\{x_k > 0\}.
\end{equation}

Then we introducing a projector to align the features from cnn feature.
During training, the feature is 50\% undergoing VQ quantization while the other is raw cnn feature.
\begin{equation}
    \mathbf{z}_{\text{TB}} = \text{TextBinarizer-Quantizer}(P(\text{VQ-Quantizer}(\mathbf{f})))
\end{equation}

where \(P\) is a lightweight transformer projector that aligns CNN feature maps to fine-grained binary representations . 
To accelerate training, we initialize both the CNN Encoder and Decoder from a pre-trained Taming-Transformer~\cite{vqgan} model and freeze these components along with the VQ quantization during training.
Following~\cite{bsq,magvit2,open-magvit2}, the training employs reconstruction, adversarial, perceptual, and commitment losses, plus an entropy penalty.

Our novel implementation offers two key advantages:
\emph{i}. \textbf{Fine-grained text feature preservation}: The binary quantization scheme captures detailed typographic characteristics essential for high-quality text rendering.
\emph{ii}. \textbf{Natural image knowledge transfer}: By leveraging pre-trained VQGAN weights and maintaining their frozen state, we retain the model's capability to handle natural images while specializing in text-focused content.
During inference, TextBinarizer operates in two modes: decoder-only for natural image generation, and full encoder-decoder for text image synthesis, providing flexibility across different use cases.

\section{\ModelName}
In this section, we unveil the Long Text Autoregressive Generation Model (\ModelName), the first framework specifically designed for generating high-quality images with extensive textual content. 
Through a novel architecture optimized for text rendering and comprehensive training on diverse text-rich datasets, \ModelName~achieves unprecedented capabilities in long-form text image generation, significantly outperforming existing models that are constrained to short phrases or single sentences.

\subsection{Architecture Overview of \ModelName}
\begin{figure}
    \centering
    \includegraphics[width=\linewidth]{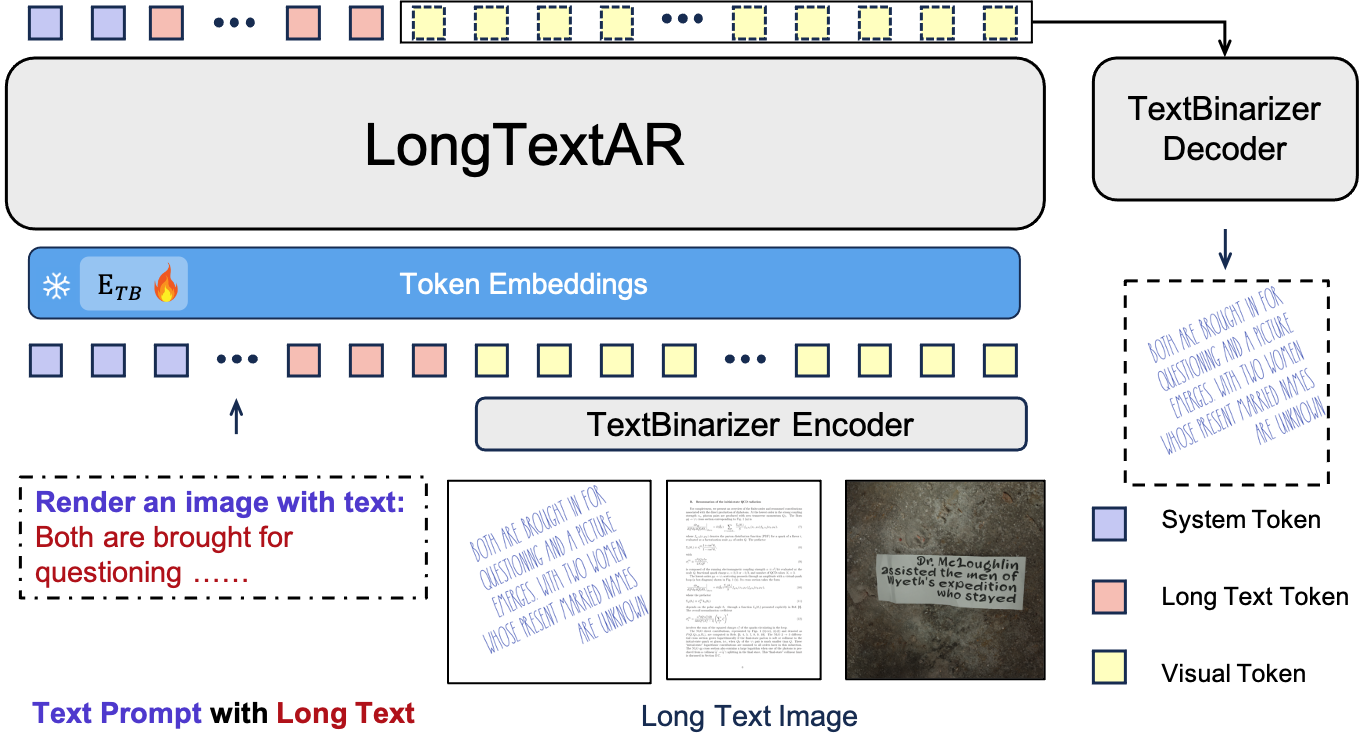}
    % \vspace{-1.5em}
    \caption{\textbf{The main pipeline of \ModelName}.
    Our trained text-focused tokenizer converts the long-text image into discrete token IDs. 
    A corresponding long-text prompt is generated, and the model is then tasked with predicting the image token IDs based on this long text prompt.
    }
    \label{fig:main_ppl}
\end{figure}

\ModelName~introduces a streamlined architecture that excels at text-to-image generation through two key components: a specialized text-focused vision tokenizer and a powerful autoregressive decoder based on Llama2~\cite{llama2}. 
As illustrated in Figure~\ref{fig:main_ppl}, our model efficiently processes long text prompts and transforms them into high-quality images with dense textual content, leveraging the ability of TextBinarizer to extract and preserve fine-grained textual features.

\paragraph{Tokenizer Integration.}
To seamlessly integrate TextBinarizer into \ModelName while maintaining optimal text processing capabilities, we implement a hybrid tokenization strategy. 
Building upon established practices in multimodal autoregressive models~\cite{showo,chameleon,lumina_mgpt}, we employ a BPE tokenizer with a vocabulary size of 65,536. 
The first \(K\) tokens are dynamically allocated to quantized representations of TextBinarizer for visual modality, while preserving the remaining slots for textual tokens.

This hybrid approach require a embedding layer that combines both visual and textual representations:
$\mathbf{E} = [\mathbf{E}_{\text{TB}}, \mathbf{E}_{\text{text}}]$,
where \(\mathbf{E}_{\text{TB}}\) represents learned embeddings of TextBinarizer for visual tokens, \(\mathbf{E}_{\text{text}}\) encompasses the embeddings for textual tokens, and $[\cdot,\cdot]$ denotes the concatenation operation to form a unified embedding space. 
To maintain the language modeling capabilities of the pre-trained AR language model, we \textit{restrict parameter updates to visual token embeddings (\(\mathbf{E}_{\text{TB}}\)) and prevent gradient propagation to text token weights in lm head layer}. 
This preserves the original language understanding of LAR model while convergencing fast.

\begin{table*}[]
    \centering
    \begin{tabular}{lll|cccc}
    \toprule
       Method  & Embed Dim & Codebook Size & FID$\downarrow$ & PSNR$\uparrow$ & SSIM$\uparrow$ & Utilization\\
    \midrule
        VQ & 512 & 8192 & 33.52 & 29.13 & 0.85 &  0.84 \\
        Taming-Transformer VQ~\cite{vqgan} & 512 & 8192 & 39.74 & 27.33 & 8.82 & 0.93 \\
        Chameleon VQ~\cite{chameleon}& 512  & 8192 & 34.63 & 29.65 & 0.86 & 0.82\\
        TextBinarizer& 512  & 8192 & 27.47 & 33.99 & 0.93 & 0.73 \\
        TextBinarizer& 512  & 65536 & 24.38 & 30.57 & 0.88 & 0.17 \\
    \bottomrule
    \end{tabular}
    % \vspace{-.5em}
    \caption{
    \textbf{TextBinarizer clearly outperform Vector Quantization on complex long-text dataset reconstruction.}
    }   \label{tab:tokenizer_comparison}
\end{table*}
\begin{figure*}
    \centering
    \includegraphics[width=\linewidth]{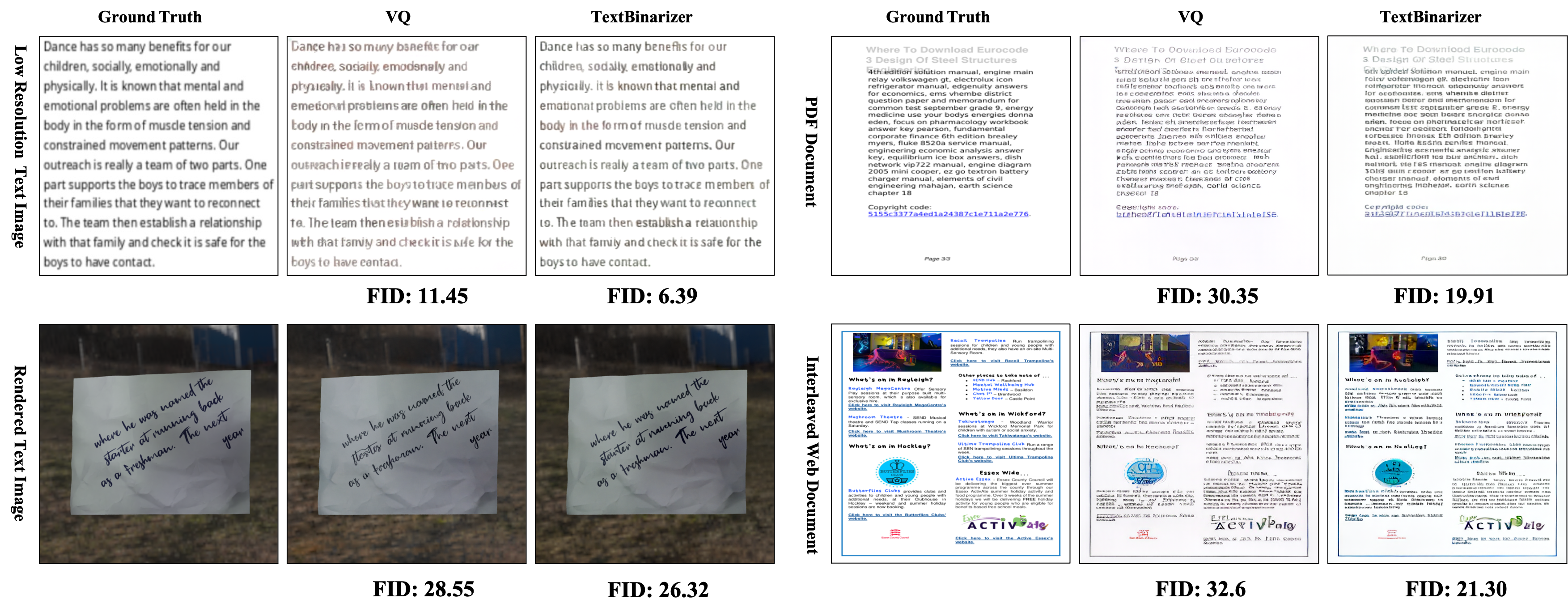}
    \vspace{-2em}
    \caption{
    \textbf{Tokenizer reconstruction comparison on data with long-text.}
    Comparing with well-trained VQ tokenizer from Chameleon~\cite{chameleon}, our text-focus tokenizer leads to better reconstruction result on detail generation for letters.
    }
    \label{fig:tokenizer_comparison}
\end{figure*}

\subsection{Elevating Text-Image Synthesis}
We employ a masked token prediction objective:
\begin{equation}
\mathcal{L}_{\text{mask}} = - \sum_{i \in \mathcal{M}} \log P(\mathbf{t}_i | \mathbf{t}_{\text{context}})
\end{equation}
where \(\mathcal{L}_{\text{mask}}\) is the loss, \(\mathcal{M}\) denotes masked positions, \(\mathbf{t}_i\) is the target text image token, and \(\mathbf{t}_{\text{context}}\) provides contextual cues include text and prompt.
All the text image tokens are masked during training.

For example, the \(\mathbf{t}_{\text{context}}\) with text variable is:
\noindent    \texttt{Generate text image with [Variables] using the following text: [$T$].}

The predicted fixed-length tokens, \(\mathbf{t}_{\text{predicted}}\), are then decoded into image via TextBinarizer:
\begin{equation}
\mathbf{x}_{\text{output}} = \text{TextBinarizer-Decoder}(\mathbf{t}_{\text{predicted}})
\end{equation}
This pipeline, depicted in Figure~\ref{fig:main_ppl}, empowers \ModelName~to fluidly synthesize text and image tokens, delivering unparalleled quality in long-text image generation.
\begin{table*}[t]
\centering
\begin{tabular}{cccccccccccc}
\toprule
Metrics & \multicolumn{2}{c}{TextDiffuser~\cite{textdiffuser}} & \multicolumn{2}{c}{AnyText~\cite{anytext}}& \multicolumn{2}{c}{TextDiffuser-2~\cite{textdiffuser2}} & \multicolumn{2}{c}{SD3.5 Large~\cite{sd3}} & \multicolumn{2}{c}{\ModelName} \\
\cmidrule(lr){2-3} \cmidrule(lr){4-5} \cmidrule(lr){6-7} \cmidrule(lr){8-9}
\cmidrule(lr){10-11}
 & Short & Long & Short & Long & Short & Long & Short & Long& Short & Long \\
\midrule
% FID${\downarrow}$ & & & & & & & & \\
CLIPScore${\uparrow}$ &31.7 & 30.6 &\underline{32.5} & 30.3 & 31.1 & 29.8 & 32.2 & \underline{30.9}& \textbf{33.3} & \textbf{31.4} \\
OCR (Accuracy)${\uparrow}$& 72.4& 38.7 & 69.5 & 34.5 & 71.3& 33.0 & \underline{73.2} & \underline{52.3}& \textbf{82.7} & \textbf{69.5} \\
OCR (F-measure)${\uparrow}$ &77.6 & 50.3& 75.49 & 51.3& 76.2 & 52.4 &\underline{78.4} &  \underline{63.5} & \bf 85.3&\bf 70.3 \\
\arrayrulecolor{gray}
\bottomrule
\end{tabular}
% \vspace{-1em}
\caption{
\textbf{Demonstration of the quantitative results for long-text rendering}. 
We split the test set into words Short (less than 10) and Long (more than 10), more details are in supplementary material.
We use the Qwen2-VL~\cite{qwen2_vl} model to recognize text from generated image.
For TextDiffuser2 model, we use prompt "A text image" with all other texts as keywords.
The best and second-best results are indicated in bold and underlined formats. \ModelName~ achieves the best results for all metrics especially for long text.}
\label{tab:performance}
\end{table*}

\section{Experiments}
% We begin by introducing the dataset and implementation details. Next, we evaluate the reconstruction capabilities of our trained text-focus tokenizer and present an ablation study on its effectiveness. 
% Following this, we empirically assess the performance of the \ModelName~ on text image generation tasks. 
% Finally, we compare our approach with existing methods and showcase potential applications.

\subsection{Datasets and Implementation Details.}
For training the TextBinarizer tokenizer, at first we construct a 11m training dataset consists of PDF data, document data, and generated text image.
For auto-regressive language model generation, we also include multiple image sources.
Mainly subsets of RenderedText~\cite{renderedtext}, Marion10M~\cite{textdiffuser}, Laion-coco~\cite{laioncoco} and AnyWords3M~\cite{anytext}.
We also include the 2 million CleanTextSynth subset from TextAtlas5M~\cite{textatlas5m}.
10\% randomly selected data used for test and the data details are given in supplementary material.

\textbf{Implementation Details.}
For TextBinarizer training, we follow the tokenizer training setting and hyperparameters in ~\cite{open-magvit2} and ~\cite{magvit2}, unless stated otherwise.
TextBinarizer is used, which eliminates the codebook embedding,  the default codebook size is $K = 2^{13}$.
For autoregressive Language Model Training, our training process uses the AdamW~\cite{adamw} optimizer, with $\beta_1$ sets to 0.9 and $\beta_2$ to 0.95, with an $\epsilon = 10^{-5}$. 
We use a linear warm-up of 4000 steps with an exponential decay schedule of the learning rate to 0.
Additionally, we apply a weight decay of 0.1 and global gradient clipping at a threshold of 1.0. 
We use a dropout of 0.1 for training stability.

\subsection{Tokenizer Impact on Reconstruction}
\label{sec:tokenizer_impact}
In this section, we evaluate our model on the test split of the tokenizer training data, which includes four types of data: (1) Text Image, (2) Rendered Text Image, (3) PPT Document, and (4) Interleaved Web Document.

We train three models on the same dataset:
1. \textbf{VQ Model (codebook size 8192):} Implementation from Taming Transformer~\cite{vqgan} and Chameleon~\cite{chameleon}.
2. \textbf{TextBinarizer Model (codebook size 8192):}.
3. \textbf{TextBinarizer Model (codebook size 65536):}.
We compare the reconstruction performance of our TextBinarizer-based text tokenizer model against the well-established VQ tokenizer from VQ-GAN~\cite{vqgan} and Chameleon~\cite{chameleon}. 
Chameleon VQ is trained on large-scale data.

The quantitative results are shown in Table~\ref{tab:tokenizer_comparison}. We observe that the TextBinarizer model demonstrates a clear improvement over the VQ model in terms of FID score, indicating better reconstruction quality. 
When we trained the VQ model on our custom data, we saw further FID improvements; however, TextBinarizer models consistently outperformed VQ models, especially as the codebook size increased. Notably, TextBinarizer-18 showed lower utilization than TextBinarizer-13, suggesting that text images do not require a very large codebook, and a dimension of 8192 is sufficient for effective reconstruction.

In addition to quantitative analysis, we also visually compare reconstruction quality across tokenizers in Figure~\ref{fig:tokenizer_comparison}. 
Our tokenizer produces noticeably better reconstructions than the Chameleon VQ-GAN when reconstructing low-resolution text images and more complex, interleaved documents. After processing these images at a resolution of 384 x 384, our TextBinarizer model achieves clearer and more accurate reconstructions, highlighting the advantages of TextBinarizer over traditional VQ approaches for text-based visual tasks.

\begin{figure}
    \centering
    \includegraphics[width=\linewidth]{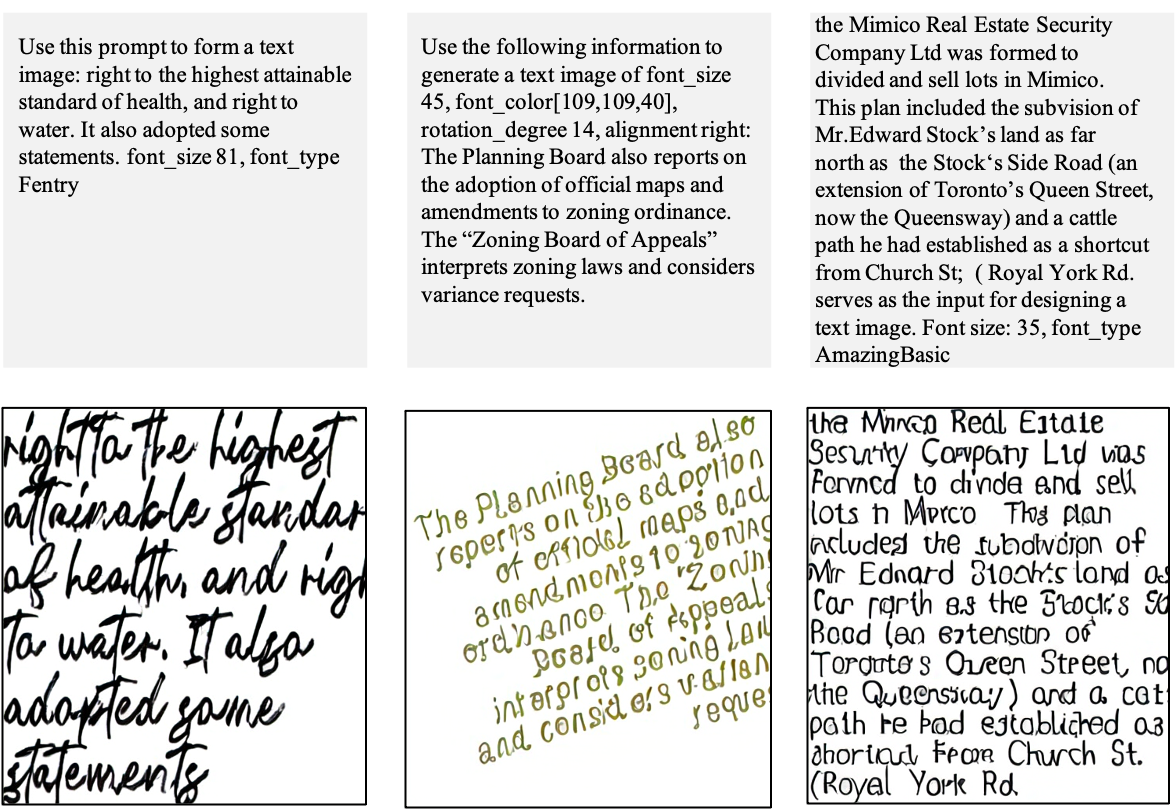}
    % \vspace{-1.5em}
    \caption{\textbf{Controllable experiment}, we modify the text font type, text color and text rotation degree, also the alignment way.}
    \label{fig:sec4_controable}
\end{figure}

\subsection{Controllable Long Text Rendering}

In this experiment, we demonstrate the ability of our model to generate text images with diverse formatting controls. 
We specify attributes such as font type, font size, rotation angle, alignment, and font color.

\begin{figure*}
    \centering
    \includegraphics[width=\linewidth]{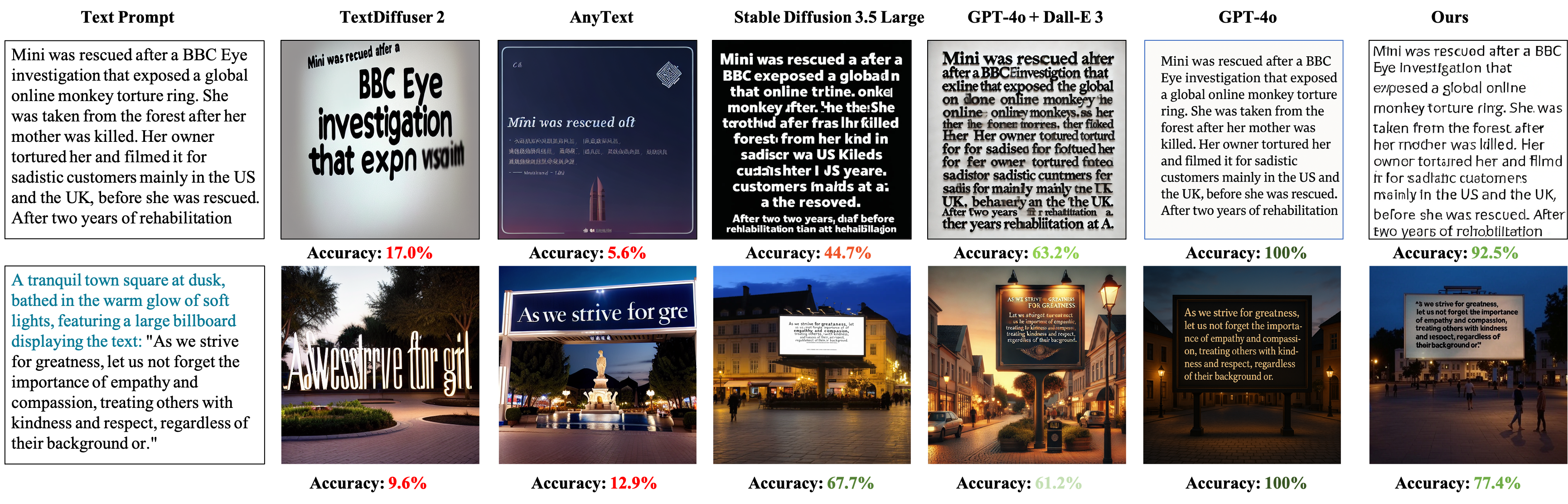}
    % \vspace{-2em}
    \caption{
    \textbf{Text-conditioned long-text image generation comparison.}
    The Stable Diffusion3.5 Large~\cite{sd3} and GPT-4o~\cite{gpt4o}+Dall-E3~\cite{dalle3} using the prompt \textit{Generate a white-background text image and the text is: [Text Prompt].
    The text is clear and large.} 
    For TextDiffuser 2~\cite{textdiffuser2} we use the prompt \textit{A text image} and input other text as tags.
    We use the Qwen2-VL~\cite{qwen2_vl} to recognize words from generated images and compute the accuracy according to the ground-truth text \textit{[Text Prompt]}.
    \textbf{The image generation capabilities of GPT4o~\cite{gpt4o}, released at the end of March 2025, have shown a huge gap over all other models, both open-source and closed-source.}
    }
\label{fig:rendering_comparison}
\end{figure*}

Figure~\ref{fig:sec4_controable} presents the results. 
Here, we showcase variations in rotation angle, font type, font size, and font color, illustrating the model’s adaptability to complex formatting requirements.
% \ModelName~ can handle various formatting cases and automatically wraps text to new lines as needed. 
Surprisingly, after exposure to over 8,000 font types during pre-training, our model can generate text in a variety of commonly used fonts with high fidelity.
For example, in the final column, we present dense text rendered with an different font. 
Our observations indicate that the majority of words are rendered accurately and are easily recognizable. Additionally, the text is automatically wrapped into different rows for better readability.
More controllable experiments are reported in the supplementary.
These evidences shows \ModelName~ can control text rendering well with common used variations.

\begin{table}[]
    \centering
    \begin{tabular}{c|x{25}x{25}x{25}x{25}}
    \toprule
    Data & \multicolumn{2}{c}{Llama2~\cite{llama2}} & \multicolumn{2}{c}{Chameleon~\cite{chameleon}}  \\
        & FID(S)$\downarrow$ & FID(N)$\downarrow$ & FID(S)$\downarrow$ & FID(N)$\downarrow$ \\
    \midrule
        Synthehic Data &28.7&73.5& \bf 27.3 & 45.6 \\
        Natural Image &69.2&82.5& 63.4& \bf 30.2  \\
        Both& 28.5 &  62.1& 28.1 & 32.4 \\
    \bottomrule
    \end{tabular}
    % \vspace{-.5em}
    \caption{
    \textbf{Comparative Analysis of Co-training with Synthetic and Natural Data and Weight Initialization Strategies.}}
    \label{tab:co_training}
\end{table}

\subsection{Comparison with Existing Models}

In this section, we evaluate long text generation on the synthetic level-1 test dataset and TextAtlasEval~\cite{textatlas5m}. 
We compare with previous state-of-the-art methods AnyText~\cite{anytext}, TextDiffuser-2~\cite{textdiffuser2}, AnyText~\cite{anytext}, SD3.5 Large~\cite{sd3}, Grok3~\cite{grok3}, Infinity~\cite{infinity} and GPT4o~\cite{gpt4o} + DALL-E 3~\cite{dalle3}.
To provide more aspect for comparison, we split the dataset into short and long subsets according to the words count less than 10 or not.
The results are presented in Tab.~\ref{tab:performance}, and our key observations are as follows:

\emph{i.} Previous state-of-the-art text rendering models, such as TextDiffuse2~\cite{textdiffuser2} and AnyText~\cite{anytext}, struggle to fully generate long text. 
The visualizations are given in Figure~\ref{fig:rendering_comparison}.
We observe that these models often produce only short text snippets and, in our experiments, fail with text containing more than ten words, consistent with findings in~\cite{textdiffuser2}.
For example, the OCR accuracy decrease from 79.5 to only 33.7.
\emph{ii.} Stable Diffusion 3.5 Large~\cite{sd3} 
demonstrate clearer better text rendering than traditional layout-dependent methods. 
% For example in.
However, as text length increases, rendering quality significantly degrades.
We observe issues such as duplicated words, inconsistencies in logic, and the generation of irrelevant words in the latter parts of the text.
\emph{iii}. Despite having only 7 billion parameters, our model demonstrates superior text generation capabilities, outperforming the 8.1B SD3.5 Large in generating coherent long text.

\begin{figure}
    \centering
    \includegraphics[width=\linewidth]{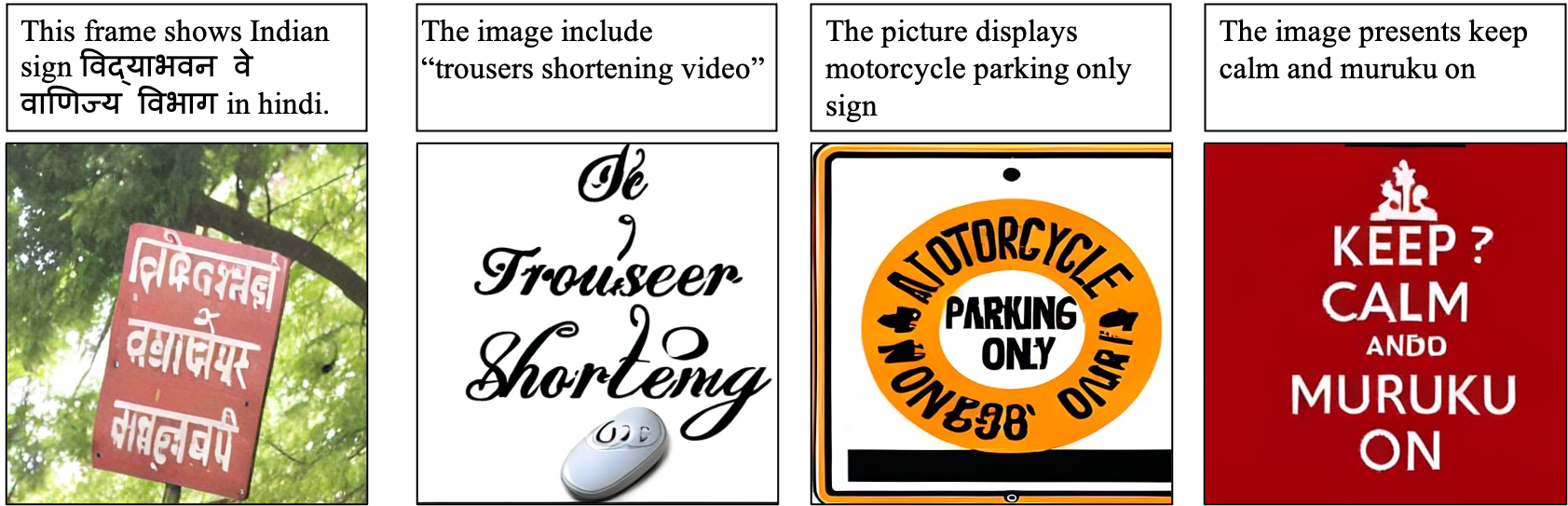}
    % \vspace{-1.5em}
    \caption{
    \textbf{Natural image text rendering examples.
    }
    }
    \label{fig:natural_image}
\end{figure}

\begin{table*}[]
    \centering
    \resizebox{\textwidth}{!}{%
    \begin{tabular}{cccccccccc|c}
    \toprule
        \multicolumn{2}{c}{Encoder} & \multicolumn{2}{c}{VQ Quant} & \multicolumn{2}{c}{Projector} & \multicolumn{2}{c}{Quant Layer 2} & \multicolumn{2}{c}{Decoder} & FID$\downarrow$ \\
        \cmidrule(r){1-2} \cmidrule(r){3-4} \cmidrule(r){5-6} \cmidrule(r){7-8} \cmidrule(r){9-10}
        Model & Frozen & Model & Frozen & Model & Frozen & Model & Frozen & Model & Frozen & \\
    \midrule
        ConvNet(256) & \checkmark & VQ & \checkmark & Single Layer FC & \checkmark & - & -& DeConvNet(256) & \checkmark & 98.44 \\ 
        ConvNet(256) & \checkmark & VQ & \checkmark & Single Layer FC &  & TextBinarizer-13 & & DeConvNet(13) &  & 79.92 \\ 
        ConvNet(256) & \checkmark & VQ & \checkmark & 3-Layer Transformer &  & TextBinarizer-13 & & DeConvNet(13) & & 62.35 \\ 
        ConvNet(256) & \checkmark & VQ & \checkmark & 3-Layer Transformer &  & TextBinarizer-18 & & DeConvNet(18) & & 58.43 \\ 
         \bottomrule
          \rowcolor{lightgray}
        ConvNet(256) &  & - & - & 3-Layer Transformer & & TextBinarizer-13 & & DeConvNet(13) & & 39.36\\ 
        \rowcolor{lightgray}
        ConvNet(13) &  & - &- & Single Layer FC &  & TextBinarizer-13 & & DeConvNet(13) & & 42.68 \\ 
    \bottomrule
    \end{tabular}%
    }
    % \vspace{-1em}
    \caption{\textbf{Training quantization and decoder experiments}. 
    The VQ-GAN weights are initialized from the Chameleon~\cite{chameleon} model, and each model is trained for 4 epochs.
    Frozen components are indicated with a \checkmark. 
    The last two rows represent the upper bounds.}
    \label{tab:w_decoder_comparison}
\end{table*}

\subsection{Ablation Study}
\subsubsection{Exploring Reuse of Pre-trained Tokenizers}
We investigate various strategies for leveraging pre-trained tokenizers. 
Specifically, we hybrid training of VQ-GAN and TextBinarizer-GAN. 
To ensure a fair comparison, all models are trained with 4 epochs and same hyperparameters.
We conduct five key experiments to assess the effectiveness of different configurations:
\emph{i}. \textbf{Pre-trained VQ Model (Baseline)}: We use the pre-trained VQ model without any fine-tuning and evaluate it directly on the validation set.
\emph{ii}. \textbf{Unfreezing Key Components}: We fine-tune the model by unfreezing the projector, TextBinarizer quantization, and decoder modules.
\emph{iii}. \textbf{Increasing Model Capacity}: We replace the single-layer projector with a 3-layer transformer to increase model capacity.
\emph{iv}. \textbf{Larger Token Embeddings}: The token embedding size is increased from 13 to 18 dimensions.
\emph{v}. \textbf{Comparison with TextBinarizer-GAN and VQGAN Encoder}: We include TextBinarizer-GAN with a $2^{13}$ codebook size and the VQGAN-Encoder for additional comparison.

The experimental results are summarized in Table~\ref{tab:w_decoder_comparison}, and the following key insights are derived:
\emph{i}. Baseline Performance: The original Chameleon VQ-GAN struggles on complex validation sets, demonstrating its limitations in handling intricate data distributions.
\emph{ii}. Unfreezing Key Components: By unfreezing the projector, TextBinarizer quantization, and decoder, the model significantly improves, reducing the FID score to 79.92.
\emph{iii}. Enhanced Projector: Replacing the single-layer projector with a 3-layer transformer yields a clear improvement, lowering the FID from 79.92 to 62.35. This highlights the importance of increasing model capacity for better tokenization.
\emph{iv.} Larger Token Embeddings: Increasing the TextBinarizer quantization codebook size further improves FID, indicating that a larger embedding space allows the model to capture more nuanced information.
\emph{v}. Unfreezing the CNN Encoder: When the CNN encoder is unfrozen, and vector quantization is removed, the model achieves even better performance than the TextBinarizer-GAN. 
This result underscores the importance of co-training the CNN encoder for optimal quantization model learning.

These findings demonstrate that careful adjustments to tokenization strategies and co-training of key components can lead to substantial performance improvements, emphasizing the critical role of tokenizers in multi-modal learning.

\subsubsection{Co-Training with Natural and Synthetic Data}

In this section, we investigate the co-training of natural images alongside generated text images. To differentiate text images from natural images, we add a placeholder, \texttt{[Natural Image]}, in the prompts. Results of this comparison are presented in Table~\ref{tab:co_training}.
We observe that the FID score for natural images remains stable at approximately consistent during pre-training, indicating that our model maintains high quality in generating realistic natural images. 
As shown in Figure~\ref{fig:natural_image}, the generated natural images retain a natural appearance. 
This demonstrates the effectiveness of our co-training approach in preserving image quality across both synthetic and natural domains.

\subsubsection{Impact of Pre-trained Weights on Generation}
In this section, we investigate the influence of pre-trained weights on the generation capabilities of the proposed model. 
Specifically, we conduct a comparative analysis using weights derived from LLaMA2~\cite{llama2} and those adapted from Chameleon~\cite{chameleon}, the latter having been pre-trained on an extensive corpus of vision-language data as an initialization strategy. 
The results, presented in Table~\ref{tab:co_training}, indicate that the Chameleon-derived weights exhibit minimal impact on text rendering performance, while significantly affecting the generation of natural images.
These findings demonstrate that \textbf{ text rendering does not necessitate strong multimodal perception from the foundation model}.
Based on these findings, we adopt the Chameleon pre-trained weights as the default initialization for the LLama2 model.

\begin{figure}
    \centering
\includegraphics[width=\linewidth]{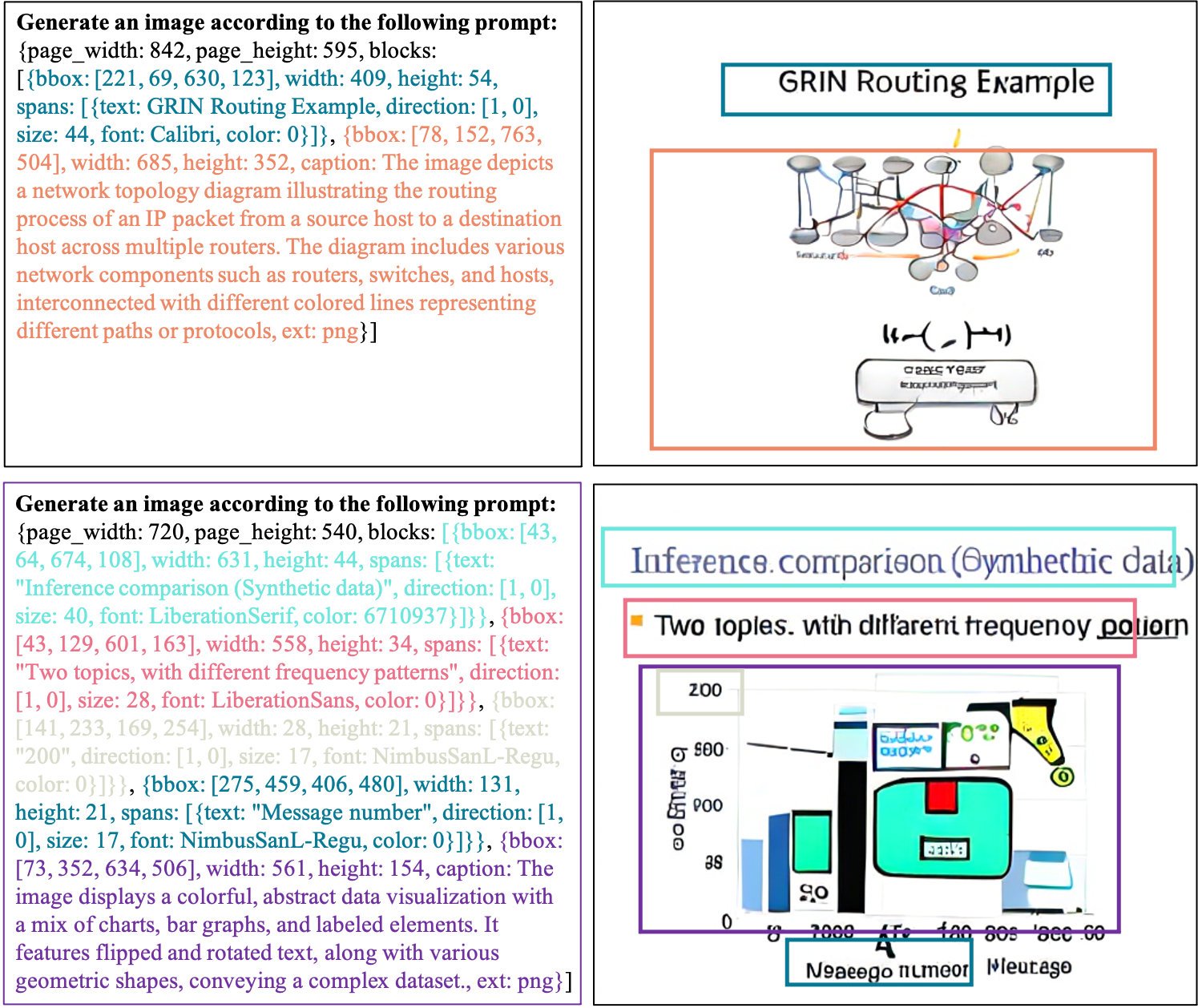}
% \vspace{-1.5em}
\caption{\textbf{\ModelName~ generates interleaved PowerPoint data from long-text prompts.} 
    The model accurately produces layouts and renders both text and images effectively. 
    Bounding boxes from the text input are shown as a reference.}

    \label{fig:ppt_application}
\end{figure}

\subsection{Potential Applications}
We leverage the AutoSlideGen~\cite{automaticslide} dataset, which consists of 5,000 PowerPoint slides, to explore the capabilities of \ModelName. 
Using PyMuPDF~\footnote{https://github.com/pymupdf/PyMuPDF}, we generate text annotations with bounding boxes and perform image captioning with the Qwen2-VL~\cite{qwen2_vl} model for images in powerpoint. \ModelName~ is fine-tuned on this reorganized dataset.

The results, shown in Figure~\ref{fig:ppt_application}, reveal that \ModelName~ excels at following positional instructions, rendering text in various formats, and generating the corresponding images. These capabilities suggest several exciting applications, such as PowerPoint editing and automated document generation. 
While the visual quality is still developing, the promising results—given the scale of the data—indicate that this approach is a practical and valuable direction for future work.
In the supplementary material, we also demonstrate that current state-of-the-art models, such as SD3.5 Large and GPT-4o with DALL-E 3, lack this capability.

\section{Conclusion and Future Directions}
In this work, we identify the weak tokenizer as a key bottleneck in autoregressive language models.
To address this, we introduce a specialized tokenizer optimized for text rendering. 
Building on this, we present \ModelName, an autoregressive transformer that, for the first time, specifically focuses on and excels at rendering long text, outperforming both traditional layout-based models and advanced large diffusion and multi-modal models.

Through comprehensive evaluations, we demonstrate robust rendering capabilities of \ModelName, particularly in generating dense text image, surpassing existing models in clarity and precision. 
We also explore controllable text rendering, mainly challenging font variations, highlighting the flexibility of our approach.
Our method opens up promising applications like powerpoint editing and interleaved document generation. 
However, the seamless integration of rendered text within natural images remains a challenging area, which we leave open for further exploration.
{
    \small
    \bibliographystyle{utils/ieeenat_fullname}
    \bibliography{main}
}

\clearpage
\setcounter{page}{1}
\maketitlesupplementary

\section{Comparison with Baseline Models Lacking Text-Focused Training}
Our method builds upon the pretrained weights of Chameleon~\cite{chameleon}. 
To evaluate the impact of text-focused training, we compare results using the baseline weights without additional fine-tuning. 
However, since the multi-modal Chameleon~\cite{chameleon} model does not provide an image generation API, we instead use Lumina-mGPT~\cite{lumina_mgpt}, an improved fine-tuned version of Chameleon optimized for high-quality image generation.

\begin{figure}
    \centering
    \includegraphics[width=\linewidth]{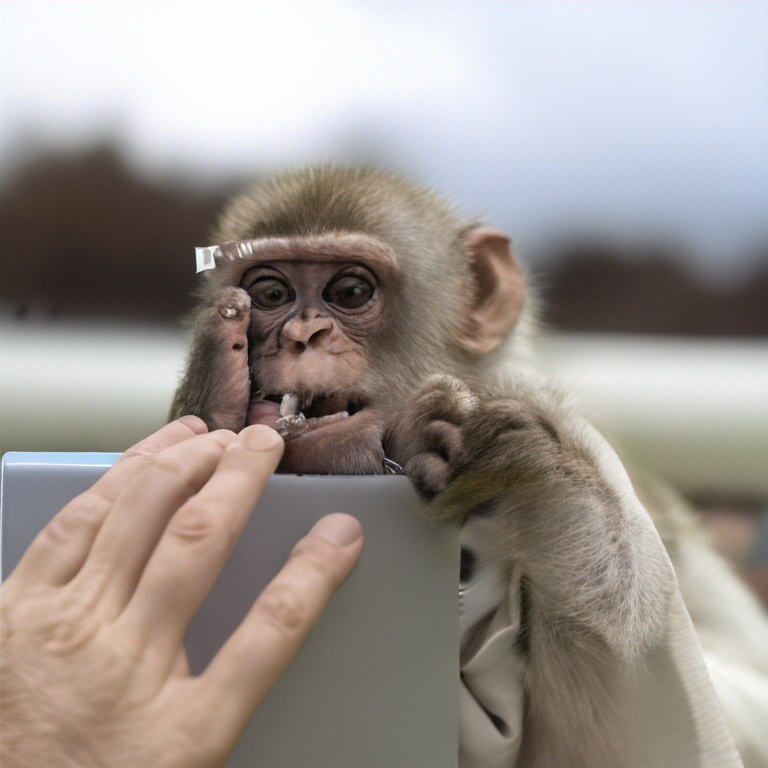}
    \caption{
    \textbf{Baseline comparison}.
    Lumina-mGPT~\cite{lumina_mgpt}, an enhanced generation model built upon Chameleon~\cite{chameleon}, demonstrates limited capability in effectively following complex textual instructions.
    }
    \label{fig:lumina_mgpt_example}
\end{figure}

\begin{figure}
    \centering
    \includegraphics[width=\linewidth]{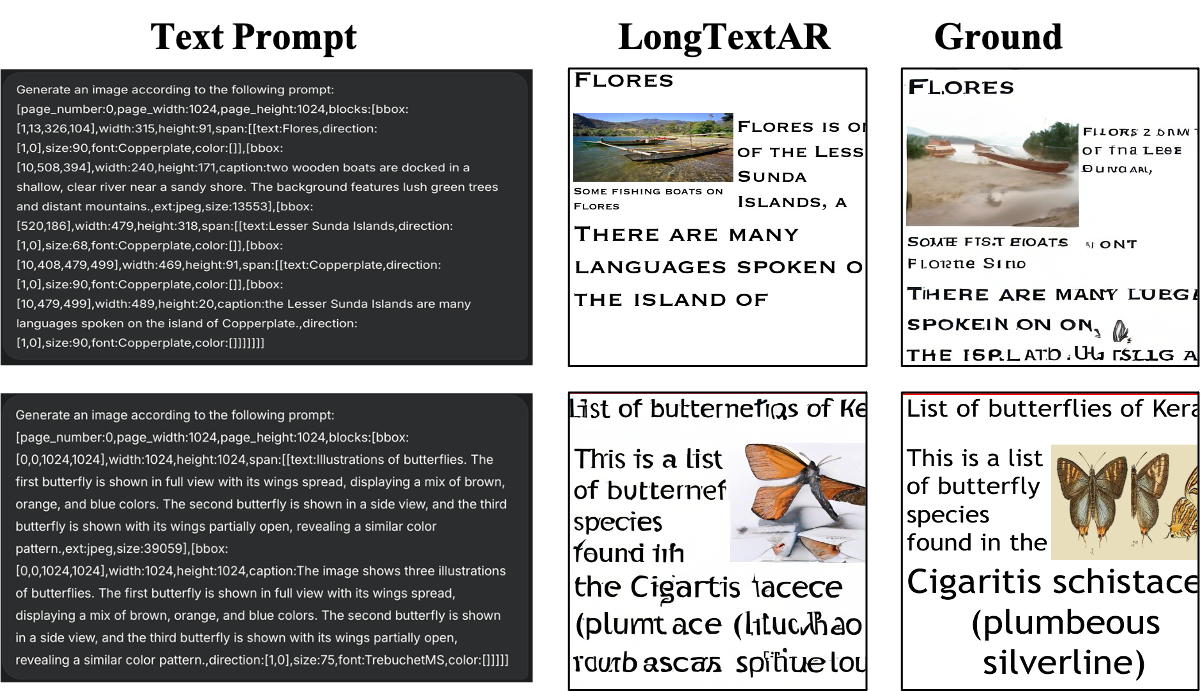}
    \caption{More interleaved data examples.}
    \label{fig:more_interleaved}
\end{figure}

The comparative results are shown in Figure~\ref{fig:lumina_mgpt_example} and Figure~\ref{fig:more_interleaved}. From these visualizations, we observe that the baseline model exhibits significant limitations, particularly in its \textbf{instruction-following ability}. For instance, while it can generate plausible images, it struggles to faithfully represent text-based prompts or instructions, leading to inconsistencies and a lack of alignment between the input text and the generated output.

\textbf{Key Observations:}
\textbf{1. Weak Instruction Alignment:} The baseline model often fails to translate detailed or specific textual instructions into corresponding visual features, demonstrating limited understanding of complex text-image relationships.
\textbf{2. Comparison with Diffusion Models:} When compared to large diffusion models and our approach, the baseline does not benefit from exposure to high-quality, text-aligned training data. This results in lower fidelity and weaker representation of fine-grained textual details in the generated images.
\textbf{3. Impact of Text-Focused Training:} Our model, leveraging additional training on text-diverse datasets, demonstrates a marked improvement in generating images that align closely with textual prompts. This highlights the importance of text-focused training for improving multi-modal instruction-following capabilities.

In conclusion, the baseline model's limitations underscore the necessity of fine-tuning on high-quality, text-aligned data to enhance instruction-following ability and ensure consistency in multi-modal image generation tasks.

\begin{figure*}
    \centering
    \includegraphics[width=\linewidth]{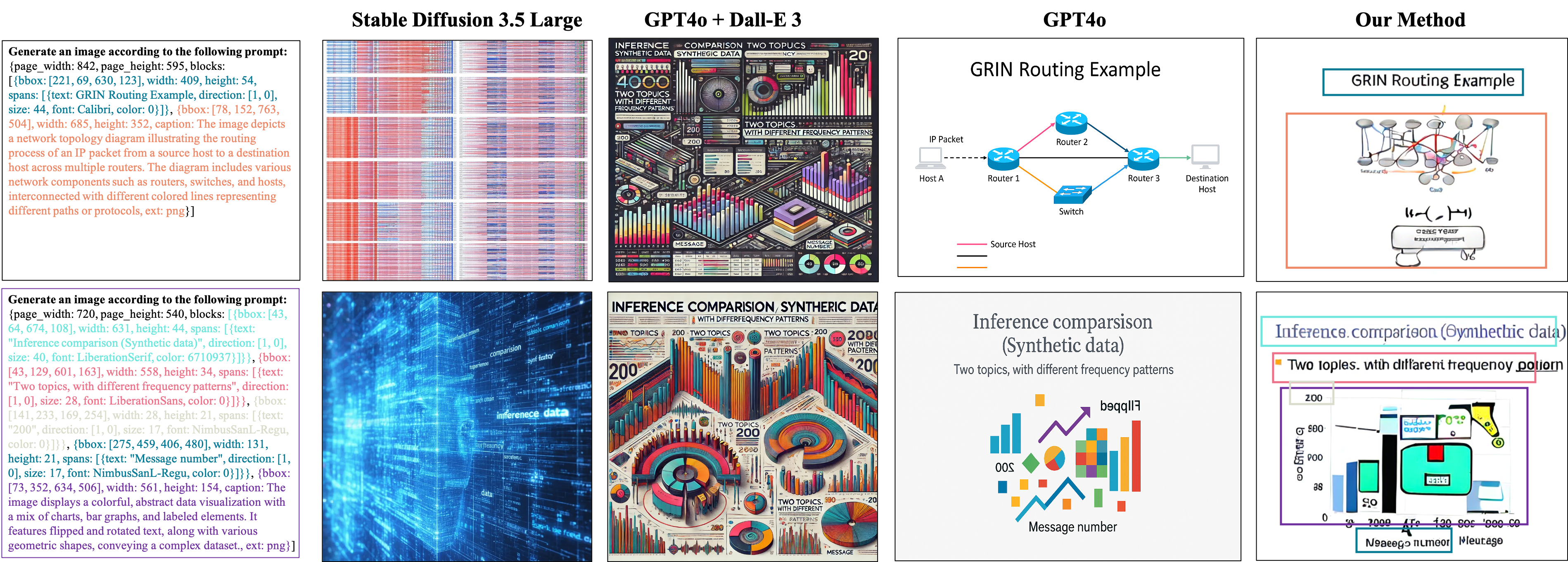}
    \caption{\textbf{The interleaved powerpoint generation comparison.}}
    \label{fig:powerpoint_comparison}
\end{figure*}

\section{Comparison with Existing Models on PowerPoint Data Generation}

Our work is designed to generate complex PowerPoint-like data from long text inputs. To evaluate its effectiveness, we test whether existing models possess similar capabilities and compare their performance.

The results of this comparison are presented in Figure~\ref{fig:powerpoint_comparison}. Below, we summarize and analyze the findings for each model:

\subsection{Performance of Existing Models}
\textbf{SD3.5 Large}~\cite{sd3}:
\begin{itemize}
    \item Demonstrates \textbf{very poor instruction-following ability}, failing to understand and process the input effectively.
    \item The generated outputs lack semantic alignment with the given instructions, rendering it unsuitable for tasks requiring structured data generation like PowerPoint slides.
\end{itemize}

\paragraph{GPT-4-O + DALL-E 3}:
\begin{itemize}
    \item Exhibits \textbf{much better instruction-following ability} than SD3.5 Large. For instance, it successfully generates a network topology diagram in the first row of the figure.
    \item Struggles with \textbf{positional accuracy} and the \textbf{correct rendering of bounding boxes}. For example, the title \textit{"GRIN Routing Examples"} is completely missing from the slide, highlighting its inability to accurately position text and visual elements.
\end{itemize}

\subsection{Performance of Our Model}
In contrast, our model demonstrates superior performance across multiple aspects:
\begin{itemize}
    \item \textbf{Accurate Layout Positioning}: 
    Our model effectively preserves the layout structure. Most elements, such as text and images, are correctly positioned within the bounding boxes, ensuring a coherent slide design.
    
    \item \textbf{Readable Text Rendering}: 
    The generated text is \textbf{highly legible}, even with variations in font styles and sizes. This reflects the model's robustness in handling diverse text formatting requirements.
    
    \item \textbf{Comprehensive Content Generation}: 
    Beyond generating high-quality images, our model accurately renders text within the visual context, combining both modalities seamlessly. This ability ensures that slides are not only visually appealing but also informationally complete.
\end{itemize}

\subsection{Analysis and Key Takeaways}
\begin{itemize}
    \item \textbf{Comparison with Existing Models}: 
    While GPT-4-O + DALL-E 3 shows promise in following instructions, its failure to capture positional details limits its usability for structured tasks like slide generation. SD3.5 Large, on the other hand, is entirely unsuitable due to its poor comprehension of inputs.
    
    \item \textbf{Strength of Our Model}: 
    By leveraging text-focused training and layout-aware design, our model bridges the gap between semantic understanding and visual generation. It ensures that both positional and textual fidelity are maintained, enabling the generation of professional-quality PowerPoint slides from long text inputs.
\end{itemize}

In conclusion, our model outperforms existing solutions in generating structured and visually coherent PowerPoint-like data, making it a reliable tool for complex data visualization tasks.

\begin{figure*}
    \centering
\includegraphics[width=\linewidth]{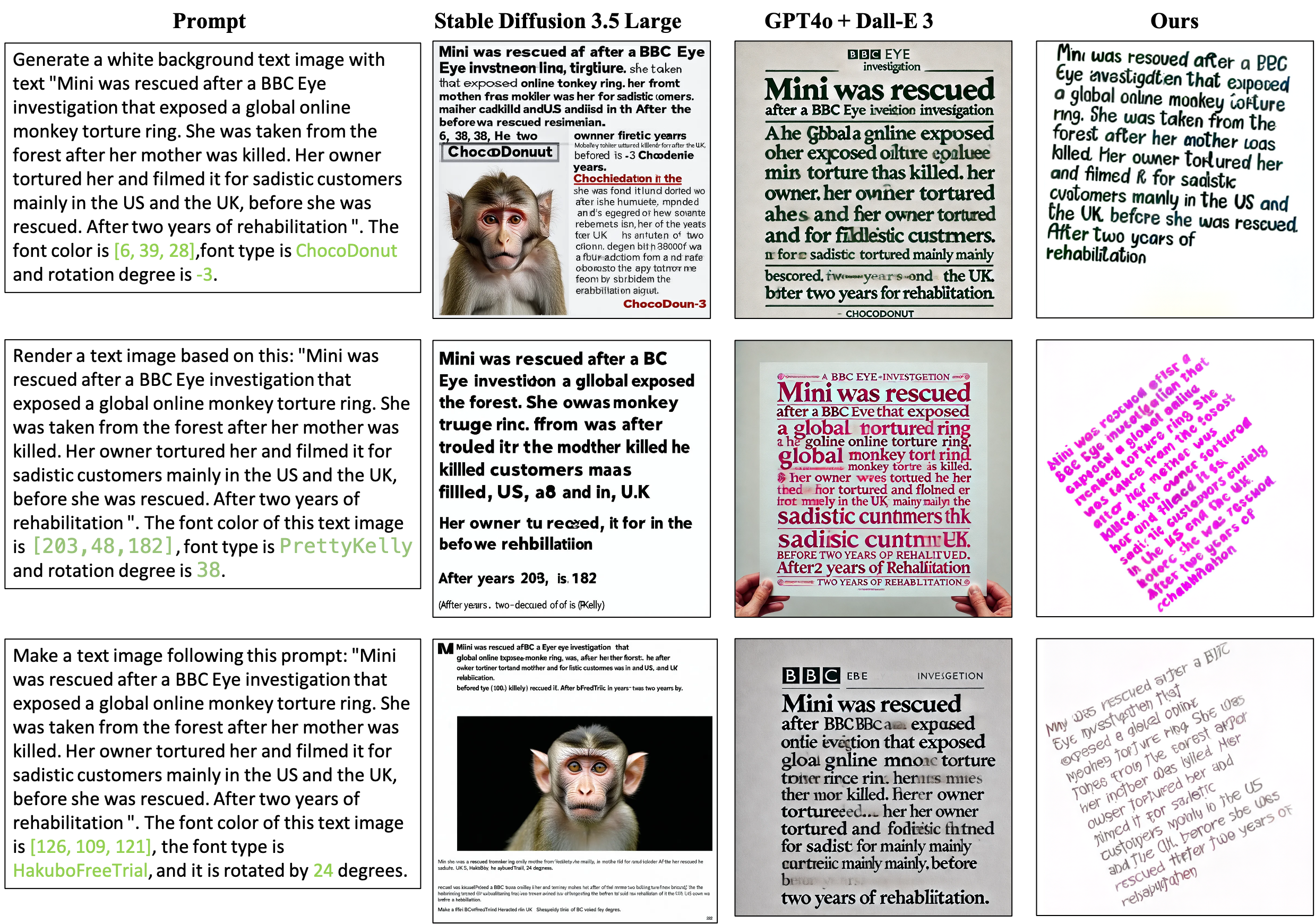}
    \caption{
    \textbf{Experiment on Controllable Variables:}
    Stable Diffusion 3.5 Large~\cite{sd3} demonstrates poor instruction-following ability, ignoring all specified variants and occasionally generating irrelevant images.  
    GPT-4-O~\cite{gpt4o}, while showing noticeably better instruction-following capability, produces outputs with low accuracy and fails to capture the specified controllable variables.
    }
    \label{fig:controlable_comparison}
\end{figure*}

\begin{figure*}
    \centering
    \includegraphics[width=\linewidth]{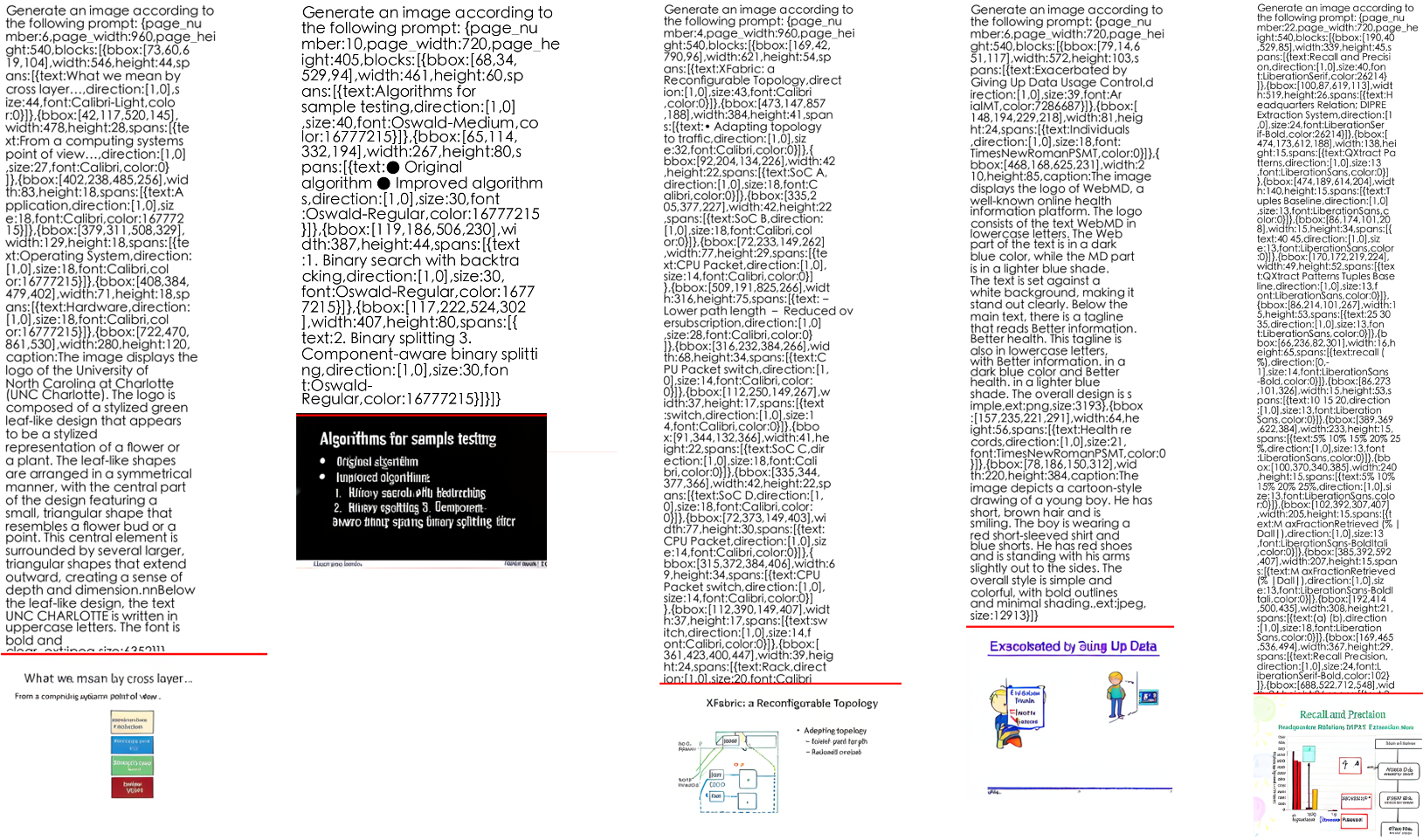}
    \caption{
    \textbf{More powerpoint generation examples.}}
    \label{fig:more_powerpoint_examples}
\end{figure*}

\section{Controllable Experiments}

In this experiment, we evaluate the ability of models to generate text based on prompts that include multiple controllable variables. The results are presented in Figure~\ref{fig:controlable_comparison}.

\subsection{Observations on Existing Models}
\textbf{Stable Diffusion 3.5 Large}~\cite{sd3}:
\begin{itemize}
    \item Often \textbf{misunderstands prompts}, extracting only a single word from the prompt to generate the image and ignoring the broader context.
    \item Attempts to improve performance with more structured prompts yielded little improvement, demonstrating its \textbf{weak instruction-following ability}, especially for complex inputs.
\end{itemize}

\textbf{GPT-4-O with DALL-E 3}~\cite{gpt4o}:
\begin{itemize}
    \item Demonstrates \textbf{better alignment} with text prompts and follows instructions more effectively than Stable Diffusion 3.5 Large.
    \item Successfully changes text colors in the middle rows of Figure~\ref{fig:controlable_comparison}, but fails to handle variables like \textbf{rotation angles}, which are absent from the outputs.
    \item Lacks diversity in text styles, as most results use \textbf{similar fonts} and exhibit minimal variation in formatting.
\end{itemize}

\subsection{Performance of Our Model}
In contrast, our model demonstrates significantly better performance:
\begin{itemize}
    \item \textbf{Comprehensive Handling of Variables}: Accurately incorporates all specified controllable variables, including font size, font color, alignment, and rotation degree, ensuring alignment with the given prompts.
    \item \textbf{Enhanced Diversity}: Generates text with substantial variation in style, such as differences in font type and size, simulating a wide range of real-world text representations.
    \item \textbf{Robust Instruction Following}: Effectively follows complex instructions, producing visually coherent and contextually accurate results.
\end{itemize}

\subsection{Analysis and Implications}
\begin{itemize}
    \item \textbf{Comparison with Existing Models}: 
    Stable Diffusion 3.5 Large struggles with prompt comprehension, while GPT-4-O with DALL-E 3, though better, still lacks diversity and the ability to handle all controllable variables.
    \item \textbf{Strength of Our Model}: 
    By explicitly focusing on controllable variables, our model bridges the gap between prompt fidelity and output diversity. This capability is valuable for synthetic dataset generation, design variations, and customized content creation.
\end{itemize}

In conclusion, our model surpasses existing solutions by accurately rendering text based on complex prompts and maintaining diversity across controllable variables, making it a versatile tool for text-to-image generation tasks.

\section{Additional PowerPoint Generation Examples}

In this experiment, we showcase more examples of PowerPoint slides generated by our model, as presented in Figure~\ref{fig:more_powerpoint_examples}. These examples highlight the model's ability to handle varying input lengths and generate diverse layouts and styles.

\subsection{Key Observations}
\begin{itemize}
    \item \textbf{Handling Variable Input Lengths}: 
    The input sequences in our examples do not have a fixed length, with some exceeding 2,000 tokens. For instance, the last example features an input sequence of 2,532 tokens. Despite this variability, our model successfully processes long sequences, demonstrating its robustness and adaptability to diverse input sizes.

    \item \textbf{Generation of Text-Only Layouts}: 
    In pure text scenarios (e.g., the second column of Figure~\ref{fig:more_powerpoint_examples}), the model produces clear and well-structured layouts. The generated slides maintain logical alignment and preserve key formatting details.

    \item \textbf{Interleaved Image and Text Data}: 
    The model also excels in generating interleaved layouts of images and text (e.g., the middle column), showcasing its capability to integrate multiple modalities seamlessly. This ability is crucial for creating complex documents that combine visual and textual elements.

    \item \textbf{Diverse Layouts and Styles}: 
    The examples reflect a variety of layouts and design styles, indicating that the model can adapt its output to suit different content types and presentation needs. This flexibility is a significant advantage for document generation tasks.
\end{itemize}

\subsection{Analysis and Limitations}
\begin{itemize}
    \item \textbf{Visual Appeal}: 
    While the generated slides demonstrate structural clarity and adaptability, the visual aesthetics still fall short of professional standards. Enhancements in style and design coherence are needed to make the outputs more visually compelling.
    
    \item \textbf{Data Challenges}: 
    The model's performance is influenced by the quality and diversity of the training data. Handling complex document structures, intricate designs, and niche use cases remains a challenge, suggesting that further research into high-quality training datasets and refinement techniques is necessary.
    
    \item \textbf{Future Directions}: 
    The results highlight the potential for future advancements in complex document generation. By addressing current limitations and improving model training with more sophisticated data, the generation of highly polished and visually appealing documents becomes feasible.
\end{itemize}

\subsection{Conclusion}
The examples demonstrate that our model can handle unfixed input lengths and generate slides with varying layouts and styles, incorporating both text-only and interleaved content effectively. Although there is room for improvement in visual aesthetics and data handling, these results lay the groundwork for further research into advanced document generation systems.

\end{document}